# Self-Compressing Neural Networks

**Szabolcs Cséfalvay**

szabolcs.csefalvay@imgtec.com

**James Imber**

james.imber@imgtec.com

## Abstract

This work focuses on reducing neural network size, which is a major driver of neural network execution time, power consumption, bandwidth, and memory footprint. A key challenge is to reduce size in a manner that can be exploited readily for efficient training and inference without the need for specialized hardware. We propose Self-Compression: a simple, general method that simultaneously achieves two goals: (1) removing redundant weights, and (2) reducing the number of bits required to represent the remaining weights. This is achieved using a generalized loss function to minimize overall network size. In our experiments we demonstrate floating point accuracy with as few as 3% of the bits and 18% of the weights remaining in the network.

## Introduction

The ongoing revolution in the capabilities of machine learning models can in large part be attributed to their increasing size. For example, the exceptional capabilities of recent state-of-the-art language models (Brown et al. 2020) have only been achieved at the expense of immense network size, slow training and execution, and high energy/carbon consumption (Lacoste et al. 2019). However, performance optimization, particularly for power- and area-efficient inference on dedicated accelerators, has been relatively neglected, which limits the deployment of powerful models on resource-limited devices (Demirci and Ferhatosmanoglu, 2021).

In this work our objective is threefold: (1) to compress networks *during* training to realize benefits in training time; (2) to reduce the size of weight and activation tensors by eliminating redundant channels; and (3) to reduce the number of bits required to represent weights. The second and third points produce a smaller network expected to execute more efficiently on devices supporting variable bit depth weight formats (Lee et al. 2019). Despite being conceptually simple, the approach we take is effective and we demonstrate high compression rates on an example classification network. We achieve the following advantages:

- Fewer weights in the final network.
- Fewer bits in the remaining parameters (depending on the target device).
- Reduced training and execution time.
- Freeing the network designer from manually optimizing architectural hyperparameters such as layer widths and bit depths.
- No requirement for special hardware to take advantage of most optimizations (e.g., no need for sparse matrix multiplication (Le Cun et al. 1989) or support for hash functions (Han et al. 2016)).

We achieve this by means of a novel quantization-aware training (QAT) scheme in which the quantization nodes are differentiable in their exponents and number of bits. This allows bit depths to be reduced simultaneously with maximizing accuracy on the task being trained for. Redundant channels are automatically detected when they reach zero bits and periodically eliminated, leading to a speedup in both training and inference due to reduced bandwidth and compute requirements.

## Related Work

Our proposed solution bridges multiple active research areas: low bit depth neural networks, QAT, and induced sparsity (particularly channel pruning).

Early contributions in the field of low bit depth neural networks showed that it is possible to achieve reasonable accuracy at very low bit depths with specialized operators (Rastegari et al. 2016, Li et al. 2016). Where specialized operators are needed, specialized inference hardware may also be required (Wang et al. 2019). The present work is designed to yield networks that may be deployed efficiently on low-bit-depth integer pipelines, as are available in many GPUs and neural network accelerators.

There exist many methods for performing QAT for network parameters. One important advance is the Straight-Through Estimator (STE) for rounding (Bengio et al. 2013), which allows gradient updates to be propagated to weights through a rounding operation during training. Other methods smooth the rounding function, using stochastic rounding (Défossez et al. 2022) or explicit smoothing (Gong et al. 2019). Importantly, Défossez et al. (2022) take QAT a step further by also learning bit depths.

The literature on induced network sparsity started with Le Cun et al. (1989). Recent related developments include methods for efficient inference of sparse networks (Demirci and Ferhatosmanoglu 2021), and induced structured sparsity such as channel pruning (He et al. 2017).

In our experiments we compare with the related method of Défossez et al. (2022), as described in more detail in the *Experiments* section below. The following differences with our method should be noted:

1. We allow bit depths to reduce to zero, eliminating some weights, instead of limiting minimum compression to 1 bit.
2. We define the quantization function in such a way that it is fully differentiable with respect to all parameters, including the number format parameters (scale/exponent and bit depth (Jacob et al. 2017)). Importantly, this turns all number format parameters into network parameters that can be trained directly as if they were weights.
3. We use the basic STE for all training instead of using pseudo-quantization noise.
4. We use a coarser grouping of weights: instead of using groups of 4, 8 or 16 weights, we group all weights in a channel, achieving greater stability and less forgetting during training. This also allows for a significant reduction in compute requirements without requiring specialized hardware by a complete elimination of channels.

## Self-Compression and Differentiable Quantization

In this paper, our experiments use a differentiable number format (eq. 1) that is shared by a group of weights, represented as signed integers with floating point exponents $e$ and bit depths $b$ (however, this is fully expected to generalize to other formats such as Q8A). Our quantization function is as follows:

$$q(x, b, e) = 2^e \lfloor \min(\max(2^{-e}x, -2^{b-1}), 2^{b-1} - 1) \rceil \quad (1)$$

Where $\lfloor \cdot \rceil$ is the rounding function which rounds to nearest integer with ties to nearest even. Since this formula is only valid for non-negative values of $b$, we constrain the range of $b$ to be greater than or equal to zero. Use of the STE to redefine the derivative of the rounding function makes it possible to optimize an objective function with respect to the quantization parameters $b$ and $e$.

The choice of rounding mode is important: when $b = 0$ the output of the $q$ function is always zero. Therefore, when a weight is represented with zero bits, it makes no contribution to the output of the network, and may be removed without changing the result. By sharing the quantization parameters across entire channels, it becomes possible to remove (prune) zero bit channels without impacting the network's output. This has the effect both of reducing the size of weight and activation tensors in the network (Figure 1), but also accelerating training over time (Figure 2) without affecting the accuracy of the final network.

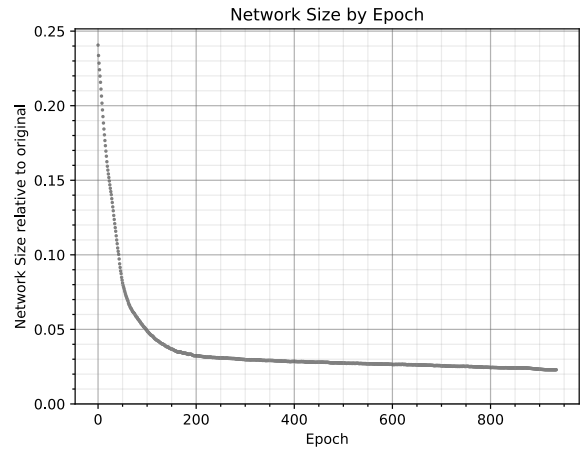

Figure 1: Using the proposed method, network size (number of bits) shrinks quickly early in training, with further reductions becoming progressively more gradual.

Reducing a network's size by removing channels has the advantage of not requiring specialized hardware to handle the reduced network. Our proposed method therefore proceeds as follows:

1. Quantizing each output channel of the weights with a single quantization parameter pair of bit depth and exponent ($b$,$e$).
2. Training the network using a loss function that maximizes accuracy on the original task whilst penalizing the number of bits used.
3. Removing network parameters (i.e. weight output channels) when the corresponding bit depths reach zero. This is also propagated to subsequent ops that consumed the removed output channel, resulting in a reduction in the size of following layers, and the removal of the corresponding input channel of a following convolution, where present.

Although the method described in this work learns to compress and eliminate channels, it is expected to generalize to other hardware-exploitable learned sparsity patterns.

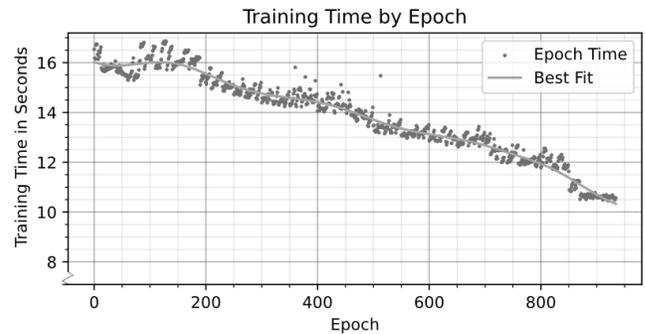

Figure 2: Training time accelerates as parameters are removed from the network.

When removing parts of a network during training the optimiser state must also be modified by removing the corresponding meta-parameters (e.g. momentum vectors) of the removed parameters.

**Optimization Objective**

In this work, it is shown that an optimization objective may be defined that improves one or more aspects of neural network performance in addition to the usual objective of reducing error on the training dataset. These aspects could include the network's size, total bandwidth consumed, number of hardware operations, power consumption, energy per inference, performance on a specific target hardware, etc. All of the above can be minimised by using bit depths as a proxy. In this work we therefore chose to minimize the number of bits, which additionally makes direct use of our proposed differentiable number format (1) for learning quantization parameters. We do this by including a new term $\gamma Q$ in the optimisation objective:

$$\Lambda(x) = \Lambda_0(x) + \gamma Q \quad (2)$$

Where $\Lambda_0$ is the original loss of the network, $\gamma$ is the compression factor (a larger $\gamma$ produces a smaller, less accurate network), and $Q$ is the average bit depth. $Q$ is defined as the sum of the sizes $z_l$ of all layers $l$, divided by the total number of weights $N$ in the starting network:

$$Q = \frac{1}{N} \sum_{l=1}^{L} z_l \quad (3)$$

The size of a layer can be expressed as the total number of bits used to represent its output channels:

$$z_l = I_l H_l W_l \sum_{i=1}^{O_l} b_l^i \quad (4)$$

Where $O_l$, $I_l$, $H_l$ and $W_l$ are the output, input, height, and width dimensions of the weight tensor of layer $l$ respectively, and $b_l^i$ is the bit depth of output channel $i$ of layer $l$. When this metric is minimized, some $b_i^l$ can reach zero. When this happens the corresponding output channel can often be removed from the network without losing accuracy.

In addition, if the output of layer $l'$ is directly used by a layer $l$, the corresponding input channel of the next layer $l$ also becomes redundant. Therefore, the compression loss may be improved by including this relationship:

$$z_l = H_w W_w \sum_{j=1}^{I} \mathbf{1}_{b_{l'}^j > 0} \sum_{i=1}^{O} b_l^i$$
$$+ H_w W_w \sum_{i=1}^{O} \mathbf{1}_{b_l^i > 0} \sum_{j=1}^{I} b_{l'}^j \quad (5)$$

Where $b_{l'}$ is the vector of bit depths used to encode the previous convolution layer's output (where present).

Once a channel can be compressed to zero bits it becomes a candidate for removal. However, removing a channel only outputting zeros could significantly change the network's output if a bias was to be added to that channel. A sudden change to the network's output can irreversibly disrupt the training, so to handle this, an $L_1$ loss is applied to biases operating on zero-bit channels to reduce them to zero. Only when the biases are reduced to zero are these output channels (and corresponding input channels from the next layer) removed, since at this point removing such a channel does not change the network's output.

A sudden change of quantisation parameters can also irreversibly degrade the network during training, which is a problem described in the next section.

**Irreversible Forgetting**

Compressing networks in this way can be challenging. We conjecture that the network is continuously trying to remove (forget) channels (or more generally groups of weights quantised by a common bit depth parameter) that are not necessary to produce a low error *at that moment* in training. However, this process could erroneously remove parts of a network that are useful, albeit not heavily used during processing of recent minibatches. For example, one might consider a network channel in the first layer trained to match horizontal lines. If multiple subsequent training batches contain no horizontal lines affecting the output, the training might determine that horizontal lines are not necessary and reduce the channel's quantization bit depth too much, and possibly to a point whereat the training can no longer relearn the feature if needed by recovering the corresponding bit depth. We will call this *irreversible forgetting*.

This phenomenon is more likely to occur deeper in the network in wider layers where more abstract (and less often-needed) features are located. We have identified ways to mitigate irreversible forgetting, including:

1. Having more weights share the same quantization parameters. Even if some of the weights in a group seem unnecessary, their encoding bit depth will stay high if other weights in the group are being used.
2. Use the Adam optimizer that adapts the learning rate when a gradient is noisy, with relatively high epsilon parameter to reduce the "acceleration" of bit depth parameters during the early phase of training.

Another factor that might affect the compression rate of the network is the error function's smoothness, but exploring this aspect is left for future work.

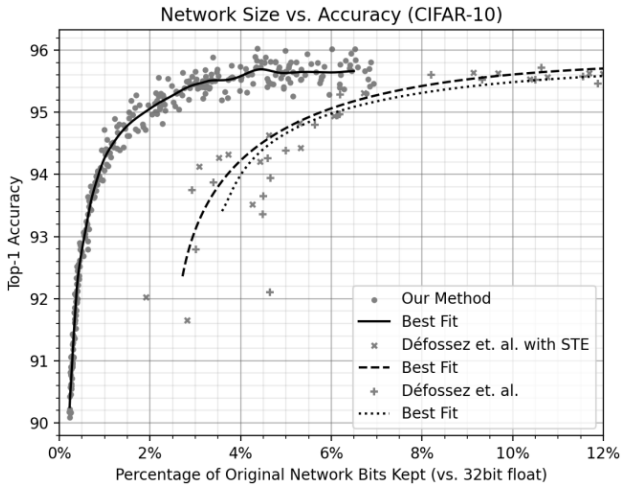

Figure 3: Top-1 accuracy on CIFAR-10 for different choices of compression factor $\gamma$. Also shown are results from the method of Défossez et al. (2022). Bit depths are determined using eq. (3) and (4) after channels of zeroes have been removed.

## Experiments

To demonstrate the proposed method, a fast-training classification network was chosen (Page 2019). This is important for being able to iterate algorithm development quickly, and to explore the tradeoff space between training time (Figure 2), network size (Figure 3), and accuracy in reasonable time.

Experiments were conducted on the CIFAR-10 dataset using the following data augmentation methods, applied in the order: (1) 4 pixel padding; (2) PyTorch AutoAugment policy for CIFAR-10; (3) random horizontal flip; (4) 32x32 random crop; (5) random erasing; and (6) normalization.

The optimizer used was Adam. For training the quantization parameters and weights we use a learning rate of 0.5 and $10^{-3}$ respectively, and an $\epsilon$ parameter of $10^{-3}$ and $10^{-5}$ respectively. A $L_2$ decay of $5 \times 10^{-4}$ was applied only to the weights. Training was run for 850 iterations, then the network was allowed to "anneal" to a final state by using PyTorch's ReduceLROnPlateau scheduler until convergence.

The same training method was used when implementing the method of Défossez et al. (2022) for fair comparison with our method.

## Results

A major advantage of Self-Compression is a parameterized trade-off between size and accuracy, in our case governed by the parameter $\gamma$ (Equation 2). The network was trained with $\gamma$ log-uniformly sampled from the interval [$10^{-3}$, $10^{-0.5}$]. As can be seen in Figure 3, this forms a locus in a plot of accuracy against final network size, wherein high values of $\gamma$ correspond to higher compression/lower accuracy. Baseline 32-bit float accuracy on this network is 95.69 ± 0.22, which we can match down to as few as 3% of the network weight bits (18% of weights) remaining.

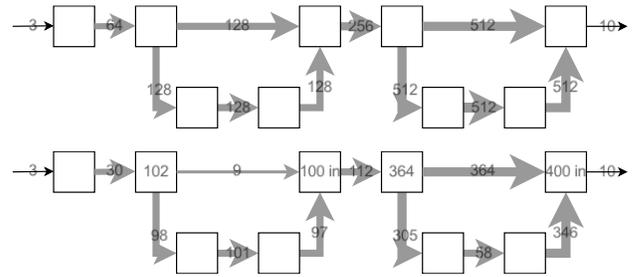

Figure 4: An overview of the number of weight channels in the example classification network before (top) and after (bottom) applying Self-Compression.

Also shown in Figure 3 are results for the method of Défossez et al. (2022) which also learns bit depths simultaneously with optimization of accuracy. Their method typically achieves floating point accuracy when the final size is above ~8% of the original number of bits. However, our proposed method maintains high accuracy at lower numbers of bits. We also note that the locus of their method is considerably noisier, which may be due to their use of a smaller weight granularity and stochastic rounding. One key difference between our proposed method and that of Défossez et al. is the form of the quantization function (STE vs. stochastic rounding). For this reason, we also include results in Figure 3 for their method using the STE instead of stochastic rounding, which results in a modest improvement in accuracy.

Figure 4 shows the number of channels before and after Self-Compression is applied with $\gamma = 0.015$. The boxes represent convolution blocks, comprising convolutions with optional batch norm and bias. The numbers on the arrows indicate number of activation channels, and the numbers on the convolution blocks represent the number of output channels. Where a summation has been performed, the number of input channels is instead noted.

## Conclusion

We have introduced Self-Compression: an efficient, conceptually simple means of learning the bit depths used to represent a network's parameters simultaneously with learning its weights, so that during training the network size is reduced simultaneously with maximizing accuracy on its task. Results on the CIFAR-10 classification task indicate that accuracy close to 32-bit floating point can be achieved with as few as 1-3% of the original bits remaining. Importantly, performance improvements are realisable on typical hardware for accelerating neural networks including

CPUs, GPUs, and neural network accelerators, without the need for specialized hardware or execution algorithms.

## Acknowledgments

Our special thanks go to Timothy Gale and Gunduz Vehbi Demirci. We would also like to thank our other colleagues at Imagination Technologies who supported this work.

## References

Bengio, Y.; Léonard N.; and Courville A. 2013. Estimating or Propagating Gradients through Stochastic Neurons for Conditional Computation. arXiv preprint. arXiv:1308.3432v1 [cs.LG]. Ithaca, NY: Cornell University Library.

Brown T.; Mann, B.; Ryder, N.;, Subbiah, M.; Kaplan, J.; Dhariwal, P.; Neelakantan, A.; Shyam, P.; Sastry, F.; Askell, A.; Agarwal, S.; Herbert-Voss, A.; Krueger, G.; Henighan, T.; Child, R.; Ramesh, A.; Ziegler, D.; Wu, J.; Winter, C.; Hesse, C.; Chen, M.; Sigler, E.; Litwin, M.; Gray, S.; Chess, B.; Clark, J.; Berner, C.; McCandlish, S.; Radford, A.; Sutskever, I.; and Amodei, D. 2020. Language Models are Few-Shot Learners. In *Advances in Neural Information Processing Systems 33 (NeurIPS)*.

Défossez, A.; Adi, Y.; and Synnaeve, G. 2022. Differentiable Model Compression via Pseudo Quantization Noise. arXiv preprint. arXiv:2104.09987v3 [stat.ML]. Ithaca, NY: Cornell University Library.

Demirci, G.; and Ferhatosmanoglu, H. 2021. Partitioning Sparse Deep Neural Networks for Scalable Training and Inference. In *Proceedings of 35$^{th}$ ACM International Conference on Supercomputing (ICS)*.

Gong, R.; Liu, X.; Jiang, S.; Li, T.; Hu, P.; Lin, J.; Yu, F.; and Yan, J. 2019. Differentiable Soft Quantization: Bridging Full-Precision and Low-Bit Neural Networks. In *Proceedings of the IEEE/CVF International Conference on Computer Vision (ICCV)*.

Han, S.; Mao, H.; and Dally, W. 2016. Deep Compression: Compressing Deep Neural Network with Pruning, Trained Quantization and Huffman Coding. In *Proceedings of the 4$^{th}$ International Conference on Learning Representations (ICLR)*.

He Y.; Zhang X.; and Sun J. 2017. Channel Pruning for Accelerating Very Deep Neural Networks. In *Proceedings of the IEEE/CVF International Conference on Computer Vision (ICCV)*.

Jacob, B.; Kligys, S.; Chen, B.; Zhu, M.; Tang, M.; Howard, A.; Adam, H.; and Kalenichenko, D. 2017. Quantization and Training of Neural Networks for Efficient Integer-Arithmetic-Only Inference. arXiv preprint. arXiv:1712.05877v1 [cs.LG]. Ithaca, NY: Cornell University Library.

Lacoste, A.; Luccioni, A.; Schmidt, V.; and Dandres, T. 2019. Quantifying the Carbon Emissions of Machine Learning. arXiv preprint. arXiv:1910.09700v2 [cs.CY]. Ithaca, NY: Cornell University Library.

Le Cun, Y.; Denker, J.; and Solla, S. 1989. Optimal Brain Damage. In *Advances in Neural Information Processing Systems 2 (NIPS)*.

Lee, J.; Kim, C.; Kang, S.; Shin, D.; Kim, S.; and Yoo; H.-J. 2019. UNPU: An Energy-Efficient Deep Neural Network Accelerator with Fully Variable Weight Bit Precision. In *IEEE Journal of Solid-State Circuits* 54(1): 173-185. doi.org/10.1109/JSSC.2018.2865489

Li F.; Zhang B.; and Liu, B. 2016. Ternary Weight Networks. arXiv preprint. arXiv:1605.04711v2 [cs.CV]. Ithaca, NY: Cornell University Library.

Page, D. 2019. cifar-10-fast. https://github.com/davidcpage/cifar10-fast. Accessed: 2022-11-03.

Rastegari, M.; Ordonez, V.; Redmon, J.; and Farhadi, A. 2016. XNOR-Net: ImageNet Classification using Binary Convolutional Neural Networks. In *Proceedings of the 14$^{th}$ European Conference on Computer Vision (ECCV)*.

Wang, E.; Davis, J.; Cheung, P.; and Constantinides, G. 2019. LUTNet: Learning FPGA Configurations for Highly Efficient Neural Network Inference. In *IEEE Transactions on Computers* 69: 1795-1808. https://www.doi.org/10.1109/TC.2020.2978817